SYNTHETIC PATIENT GENERATION:

Deep learning to generate new patient records

Ally Salim Jr

Inspired Ideas LLC - Dr. Elsa

**Abstract**

Artificial Intelligence in healthcare is a new and exciting frontier and the possibilities are endless. With deep learning approaches beating human performances in many areas, the logical next step is to attempt their application in the health space. For these and other Machine Learning approaches to produce good results and have their potential realized, the need for, and importance of, large amounts of accurate data is second to none. This is a challenge faced by many industries and more so in the healthcare space. We present an approach of using Variational Autoencoders (VAE's) as an approach to generating more data for training deeper networks, as well as uncovering underlying patterns in diagnoses and the patients suffering from them. By training a VAE, on available data, it was able to learn the latent distribution of the patient features given the diagnosis. It is then possible, after training, to sample from the learnt latent distribution to generate new accurate patient records given the patient diagnosis.

Synthetic Patient Generation: Deep learning to generate new patient records

## 1. Introduction

While researching the use of Deep Learning in assisting health care workers to diagnose and treat patients, the data collection process turned out to be unimaginably cumbersome and infinite. With challenges from availability of relevant accurate data, to availability of digital data, to cryptic doctors handwritings, resulting in long data entry times, long periods of searching of relevant data, minor mistakes due to the difficult to understand handwritings and not to mention the high cost of acquiring and digitizing the data.

Given the challenges, we turned to an attempt at generating synthetic data that we can then use to learn more and improve our performance in the wild. After experimenting with a few approaches like the gaussian mixture models and adversarial networks, we settled on Variational Autoencoders as our generative model for its simplicity and performance.

Variational Autoencoders, introduced by Kingma and Welling in 2013, are a deep learning technique for learning latent distributions in data [1]. Their uses are vast, ranging from image generation [2], to sentence interpolation [3] and are particularly useful in learning hidden concepts and uncovering relationships in large quantities of unlabeled data.

The training is done with patient records of one diagnosis at a time, where the encoder part of the network encodes the data into smaller dimensions by having fewer nodes at the output than at the input, also known as a *'bottleneck'*. The encoder outputs parameters to $q\theta(z|x)$ which is a gaussian probability density. The decoder will take the output parameters from the encoder and attempt to re-create the original data fed into the encoder, resulting in a *slightly fuzzy* copy of the original record.

## 2. Methods
### 2.1 Datasets

The dataset used is privately owned by Inspired Ideas LLC and was collected from partner hospitals in the regions of Arusha and Dar es Salaam, Tanzania. Spanning across 9 years, the dataset is a series of records of patient visits, containing the patient symptoms, age, gender, time of year, differential diagnosis, tests ordered, test results, diagnosis, and treatments.

Although the dataset covers many of the common ailments and how to deal with them, it is by no means comprehensive as it has very few and sometimes no records of the extremely rare conditions. This might be an advantage in that there is no need in learning the rare things that will not be needed, but a disadvantage in generalizability to other regions of the continent, or even the other end of the country.

**2.2 Variational Autoencoders**

Variational Autoencoders (VAE), are a type of semi-supervised/self supervised learning neural network architecture that are a part of the Autoencoder family. The VAE's can be seen as neural generative models that learn to represent a dataset using a gaussian distribution by first *encoding* the input data into a lower dimensional space (a gaussian distribution density) and then *decoding* the a sample from this distribution back to the original input. In other words, during training, this type of neural network attempts to recreate the input given severe limitations (fewer nodes in hidden layers). See figures 2.2 and 2.3 for a visual representation of this information.

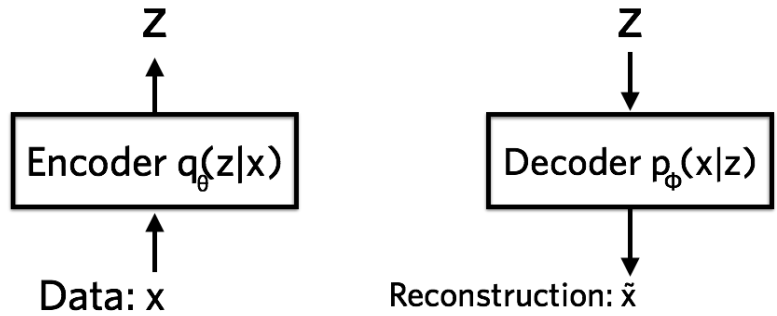

*Figure 2.2: Visual structure of autoencoder [4].*

While Variational Autoencoders can been used to compress data into lower dimensional data, they can also be used as powerful generative models. By sampling from the distribution *qθ(z|x)*, we can generate completely new data that is *inferred* from all the examples it has seen before. Bayes theorem says:

$$p(z|x) = \frac{p(x|z)p(z)}{p(x)}.$$

Where the denominator *p(x)* is known as the evidence and can be calculated by:

$$p(x) = \int p(x|z)p(z)dz$$

Unfortunately calculating this requires exponential time, therefore we need to approximate the posterior distribution with $q_\lambda(z|x)$ where $\lambda x_i = (\mu x_i, \sigma x_i^2))$.

**2.2.1 Measuring VAE performance**

Conventional loss functions fall short due to the unique nature of Variational Autoencoders. For this we use the *Kullback–Leibler divergence* introduced in 1951 by S. Kullback and R. Leibler in 1951 [5].

The *Kullback–Leibler divergence*, also known as *relative entropy of P with respect to Q,* is a good way to measure how much information is lost when approximating *p* using *q*, and is expressed as:

$$KL(q_\lambda(z|x)||p(z|x)) =$$
$$\mathbf{E}_q[\log q_\lambda(z|x)] - \mathbf{E}_q[\log p(x,z)] + \log p(x)$$

Notice that the difficult to calculate *p(x)* is still there, to get rid of it we introduce the Evidence Lower BOund (ELBO) concept:

$$ELBO_i(\lambda) = E_{q_\lambda(z|x_i)}[\log p(x_i|z)] - KL(q_\lambda(z|x_i)||p(z)).$$

Since in Variational Autoencoders there are only latent variable, we can further decompose the ELBO into a sum where each term depends on one data point. Where the ELBO for a single point is:

$$\log p(x) = ELBO(\lambda) + KL(q_\lambda(z|x)||p(z|x))$$

Finally, we parametrize the approximate posterior $q_\theta(z|x,\lambda)$ with an encoder and parametrize the likelihood $p(x|z)$ with a decoder. The encoders and decoders have parameters θ and ϕ respectively, typically weights and biases in neural network speak. Including the parameters of the inference and generative networks, we can now write our ELBO as:

$$ELBO_i(\theta, \phi) = E_{q_\theta(z|x_i)}[\log p_\phi(x_i|z)] - KL(q_\theta(z|x_i)||p(z)).$$

It should be clear by now that to achieve *Variational Inference*, we aim to maximize the *ELBO* with respect to the variational parameters $\lambda$.

In summary, variational autoencoders are able to define the probability distribution of the inputs as a latent variable space. We can then sample from this distribution and create new and never before seen patient records (given a diagnosis).

**2.3 Network Architecture**

As seen in figure 2.3, the network used during the experiments has a total of 5 layers (counting the input, and output layers).

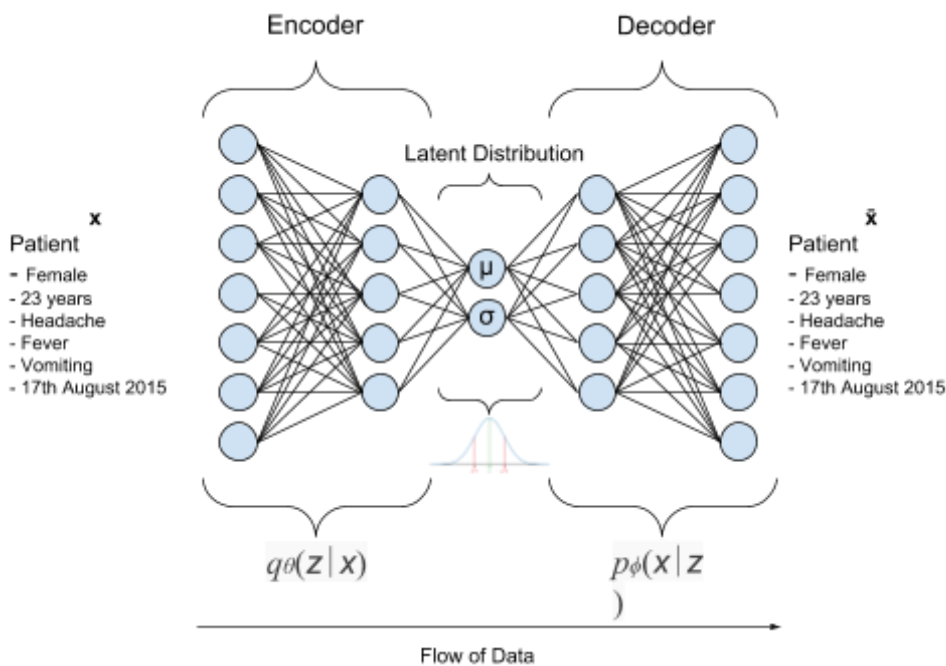

*Figure 2.3: Complete Network architecture used.*

**2.4 Training**

Table 2.4.a shows a visualization, in the form of a PCA, illustrating how the network learns to create realistic synthetic patient data starting from a very poor generation at epoch 0 to a very accurate generation in only 90 epochs. During training it is important to experiment with the number of layers (more for larger datasets and less for smaller datasets) and number of neurons per layer (ideally larger layers at the input and at the output than in the hidden layers).

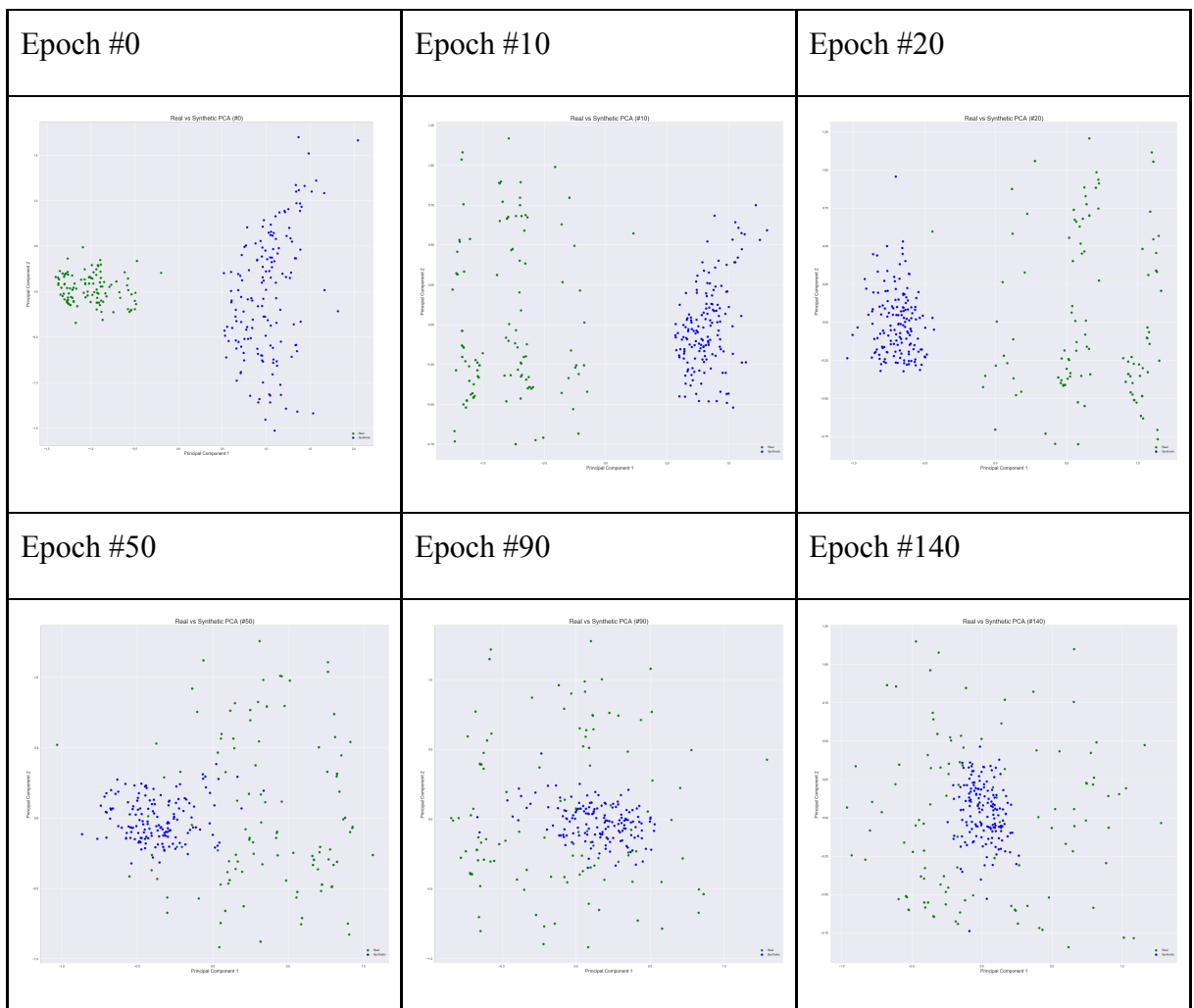

*Table 2.4.a: PCA of real patients (green) and synthetic patients (blue) generated by the network after various stages of training. Notice the large difference between the types of patients at the beginning of the training and how, progressively, the network learns to create synthetic patients that are similar to real patients.*

## 3. Results

The results are impressive for such a small dataset, generating results that are in some cases indistinguishable from real data. This is because the generative model has learnt the latent representation of the data and is able to sample from the distribution to create variations of the data. Tables 3.a, 3.b, and 3.c show sample generations given a diagnosis.

**Gastroenteritis**

| Patient No. | Gender | Age (yrs) | Month | Symptoms |
|---|---|---|---|---|
| 1 | Female | 28.3 | April | diarrhea, fever, vomiting |
| 2 | Male | 40.7 | August | abdominal pain, diarrhea, fever, vomiting |
| 3 | Female | 31 | April | abdominal pain, diarrhea, fever, vomiting |
| 4 | Male | 0.5 | July | cough, diarrhea, fever, vomiting |
| 5 | Female | 29 | April | abdominal pain, body weakness, diarrhea, fever, vomiting |

*Table 3.a. Synthetic patients with Gastroenteritis.*

**Pneumonia**

| Patient No. | Gender | Age (yrs) | Month | Symptoms |
|---|---|---|---|---|
| 1 | Female | 13.9 | April | cough, difficulty breathing, fever |
| 2 | Male | 29.7 | November | cough, coughing up blood, fever, general body malaise |
| 3 | Male | 0.5 | November | chest pains, cough, difficulty breathing, fever |
| 4 | Female | 15.8 | January | cough, difficulty breathing, fever |
| 5 | Female | 0.5 | November | cough |

*Table 3.a. Synthetic patients with Pneumonia.*

**Malaria**

| Patient No. | Gender | Age (yrs) | Month | Symptoms |
|---|---|---|---|---|
| 1 | Female | 39.1 | April | body weakness, fever, headaches, vomiting |
| 2 | Male | 47.0 | April | body weakness, fever, headaches, joint pain |
| 3 | Female | 33.5 | February | body pain, fever, headaches |
| 4 | Male | 84.4 | April | body weakness, convulsion, fever, sleepiness |
| 5 | Female | 29.8 | February | fever, headaches |

*Table 3.a. Synthetic patients with Malaria.*

## 3.1 Results Evaluation

To test the accuracy of the generated synthetic patients, we combined the synthetic patients with the real patient data obtained from a local hospital, and asked practicing Medical Doctors to see if they can identify which data is synthetic. The results are shown below:

| Metric | Percentage (%) |
|---|---|
| Synthetic identified as synthetic | 20.0% |
| Real identified as synthetic | 23.3% |
| Synthetic identified as real | 80.0% |

*Table 3.1.a. Table showing how how the doctors performed at identifying the synthetically generated patients. From this it is clear that the algorithm has learnt how to generate realistic data that look real to even the trained professional.*

**Outcome 1**

The Variational Autoencoder can learn to model a single diagnosis after only a few epochs on a dataset of ~ 150 data points per diagnosis. Although extremely small for deep learning standards, more data yields exponentially better results. The results, after PCA, appear indistinguishable from real records obtained from the dataset.

**Outcome 2**

Interesting observations, that warrant further research, were made as the model tended to produce results with patterns in the patient demographics and time of year that could tend towards understanding which patient demographics are at risk of certain illnesses and at what times of the year.

## 4. Conclusion

Variational Autoencoders (VAE) are powerful generative models, with roots deep in statistics and probability, whose use cases span and potential span across many industries. They are part of the Autoencoder (AE) family of *self-supervised learning* neural networks that learn by attempting to recreate the given input while under constraints such as smaller hidden layers and other regularization techniques. Variational Autoencoders achieve this by learning the latent distribution of the data, and the sampling from this distribution to re-create the input. During inference, the neural network can simply sample from the distribution and create new points that were not seen during training.

We presented an approach to patient synthesis as a solution to both the shortage in quality and general availability of patient data in Tanzania, and open avenues for disease research through using VAE's in uncovering hidden patterns and a deeper understanding of diseases and the patients suffering from them.

## 5. Acknowledgments

I am thankful to Megan E. Allen for her support in the collection and processing of the data, and the writing of the paper. Also thanks to Dr. Hassan Mshinda for creating an environment that allowed us to learn and start experimenting on this as an alternative solution.

SYNTHETIC PATIENT GENERATION	11